\documentclass[hidelinks,10pt,conference]{IEEEtran}

\usepackage[utf8]{inputenc}
\usepackage[T1]{fontenc}

\usepackage{acronym}
\usepackage{cite}
\usepackage{dblfloatfix}
\usepackage{graphicx}
\usepackage{multirow}
\usepackage{xcolor}

\usepackage{enumitem}

\usepackage{hyperref}
\usepackage[capitalise]{cleveref}

\setlist{nosep}

\crefname{paragraph}{Section}{Sections}

\title{Safety Margins for Reinforcement Learning}
\author{%
\IEEEauthorblockN{Alexander Grushin}%
    \IEEEauthorblockA{\small Galois, Inc.\\
        \texttt{agrushin@galois.com}}
\and
\IEEEauthorblockN{Walt Woods}%
    \IEEEauthorblockA{\small Galois, Inc.\\
        \texttt{waltw@galois.com}}
\and
\IEEEauthorblockN{Alvaro Velasquez}%
    \IEEEauthorblockA{\small  University of Colorado Boulder\\
        \texttt{alvaro.velasquez@colorado.edu}}
\and
\IEEEauthorblockN{Simon Khan}%
    \IEEEauthorblockA{\small Air Force Research Laboratory\\
        \texttt{simon.khan@us.af.mil}}
}
\date{\today}

\usepackage[pscoord]{eso-pic}

\begin{document}
\bstctlcite{IEEEexample:BSTcontrol}

\maketitle
\thispagestyle{plain}
\pagestyle{plain}
\begin{abstract}
Any autonomous controller will be unsafe in some situations. The ability to quantitatively identify {\em when} these unsafe situations are about to occur is crucial for drawing timely human oversight in, e.g., freight transportation applications. In this work, we demonstrate that the {\em true criticality} of an agent's situation can be robustly defined as the mean reduction in reward given some number of random actions. Proxy criticality metrics that are computable in real-time (i.e., without actually simulating the effects of random actions) can be compared to the true criticality, and we show how to leverage these proxy metrics to generate {\em safety margins}, which directly tie the consequences of potentially incorrect actions to an anticipated loss in overall performance. We evaluate our approach on learned policies from APE-X and A3C within an Atari environment, and demonstrate how safety margins decrease as agents approach failure states. The integration of safety margins into programs for monitoring deployed agents allows for the real-time identification of potentially catastrophic situations.
\end{abstract}

\section{Introduction}

Broader adoption of autonomous controllers for real-world applications relies on the ability to ensure that any benefits of automation come without unacceptable costs; for example, in freight transportation, accidents result not only in damages to the autonomous vehicle, but also loss of cargo and loss of life. There is significant other work in the field on improving the reliability of these controllers. Instead, we seek to understand and quantify when a controller might be on the brink of disaster, in order to raise an alert for timely human oversight.

During the past several years, {\it criticality metrics} have been developed for gauging the importance of any given point in time to an agent's overall success \cite{lin2017,huang2018,spielberg2018,guo2021}. These metrics are typically validated by injecting random or adversarially worst-case actions at times that have the top-N largest metric values; if this results in a large measured reduction in reward, then the metric is considered to be more accurate \cite{lin2017,guo2021}. Such an evaluation approach lacks ground truth (i.e., the {\it true criticality} at any specific time is unknown), and is liable to miss false negative errors, where the metric has a low value, but true criticality is high and the agent is in imminent danger.  Thus, the accuracy of existing metrics is not well-established; furthermore, it can be unclear what a given metric value implies in terms of potential consequences to the agent.

Instead, we re-label these metrics as {\it proxy} criticality measurements, and introduce a definition for true criticality at some point in time $t$ as the expected reduction in reward when an agent executes a sequence of $n$ consecutive random actions (beginning at $t$), rather than the actions suggested by its policy; this is a modification of a definition given in \cite{huang2018}. We present an algorithm for accurately approximating the expected value as a mean reward reduction, in a tractable way (albeit not in real-time). By analyzing the relationship between proxy and true criticality, we can identify both false positives and false negatives. Crucially, we show that even a noisy relationship between proxy and true criticality can provide actionable information. Specifically, we define the {\it safety margin} at some time $t$ as the maximum number of random actions which, if executed beginning at $t$, have only an $\alpha$ chance of impairing agent performance more than some tolerance $\zeta$, defined in the discounted reward space of the application. Intuitively, this can be illustrated in the game of Pong: a ``safe'' policy that keeps the ball centered on the agent's paddle can afford mistakes, whereas an agent that keeps the ball near the edge of its paddle may lose a point if it makes a mistake. For proxy criticality metrics that can be computed in real-time, safety margins result in a lookup table that can be consulted anytime, allowing autonomous systems to automatically flag themselves as needing human oversight in critical situations.

\section{Results}

\begin{table}[b]
    \centering
    \caption{Predictive capabilities of safety margins}
    \label{cai:tab:results:safety}
    \begin{tabular}{|c|c|c|c|}
    \hline
    $\zeta$ & algorithm & steps before death & safety margin \\\hline
    \multirow{6}{*}{0.5} & \multirow{4}{*}{APE-X} & 1 & $0.00 \pm 0.00$ \\
    & & 2 & $0.50 \pm 0.87$ \\
    & & 4 & $0.99 \pm 0.73$ \\
    & & average & $0.97 \pm 0.86$ \\\cline{2-4}
    & \multirow{4}{*}{A3C} & 1 & $0.53 \pm 0.50$ \\
    & & 2 & $1.00 \pm 0.00$ \\
    & & 4 & $0.82 \pm 0.38$ \\
    & & average & $3.09 \pm 2.68$ \\\hline
    \multirow{6}{*}{1.0} & \multirow{4}{*}{APE-X} & 1 & $2.00 \pm 0.00$ \\
    & & 2 & $2.50 \pm 0.87$ \\
    & & 4 & $4.50 \pm 2.23$ \\
    & & average & $4.72 \pm 2.50$ \\\cline{2-4}
    & \multirow{4}{*}{A3C} & 1 & $2.48 \pm 1.75$ \\
    & & 2 & $4.00 \pm 0.00$ \\
    & & 4 & $3.42 \pm 2.66$ \\
    & & average & $9.20 \pm 5.24$ \\
    \hline
    \end{tabular}
\end{table}

\begin{figure}[t]
    \centering
    \includegraphics[width=.99\linewidth]
    {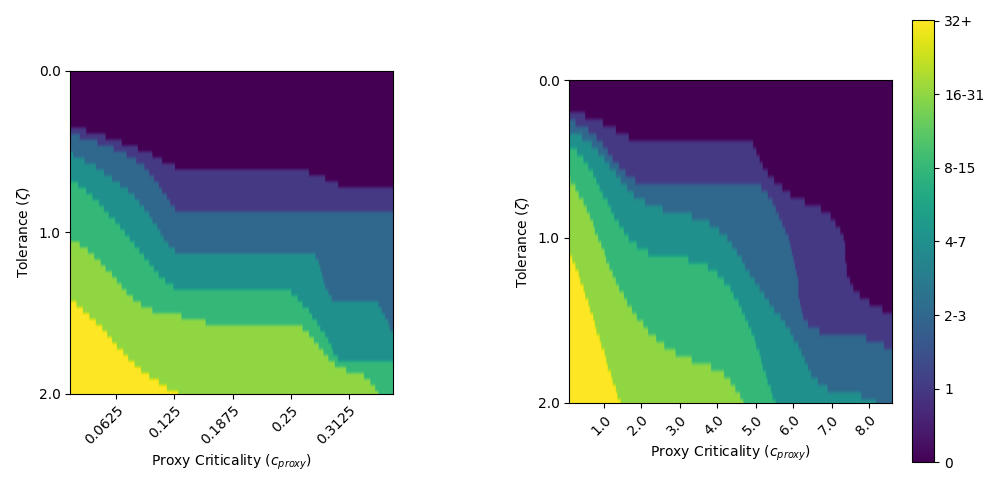}
    \caption{Lookup tables (drawn as heatmaps) displaying safety margins in the proxy criticality metric vs. tolerance space (loss in reward), for APE-X (left) and A3C (right).}
    \label{cai:fig:results:safety}
\end{figure}

We ran experiments with the Atari game BeamRider, where players pilot a ship and attempt to destroy enemy ships. Using a proxy criticality metric adapted from \cite{lin2017}, which takes the maximum predicted Q value for APE-X or action log likelihood for A3C, and subtracts the minimum such quantity, we constructed the safety margin tables shown in \cref{cai:fig:results:safety}. These lookup tables provide approximately $1-\alpha$ confidence that a given, tolerable loss in performance (Y-axis) will not be exceeded when an agent in a situation corresponding to a given proxy criticality metric (X-axis) makes a number of potential mistakes through random actions (color), with $\alpha = 0.05$.

For these safety margins to have real-world value, they must correctly indicate the proximity of the agent to a critical situation. When the agent's ship is destroyed, the agent's policy, $\pi$, was suboptimal -- that is, the agent did not know how to perform better. Nonetheless, in \cref{cai:tab:results:safety}, safety margins tend to decrease as the agent nears its own destruction, and are much lower than the mean safety margins (averaged over all times in all episodes). This indicates the ability of our defined safety margins to correctly identify when the agent would benefit from human oversight.  We further note that for APE-X, in 22\% of cases, the proxy criticality value just before the agent's death was within the top 5\% of proxy criticality values sampled across all times in all episodes, and for A3C, the corresponding proportion was 47\%. This suggests that even if the human intervenes only when proxy criticality is very high, safety can potentially be improved; further improvements might be yielded with more accurate proxy metrics.

While the $1 - \alpha$ confidence aspect of safety margins addresses false negatives, we also note that the co-occurrence of low proxy criticality values with high true criticality can be studied through density plots created in the process of computing safety margins, as in \cref{cai:fig:noise}. When a point in time with a low proxy metric (left X-axis) can have a significant reduction in reward triggered (bottom Y-axis), users can decide if, e.g., an $\alpha = 0.01$ confidence level would be more appropriate to avoid the potential dangers of false negative situations.

\section{Methods}\label{cai:sec:methods}

\begin{figure}[t]
    \centering
    \includegraphics[clip, width=\linewidth]
    {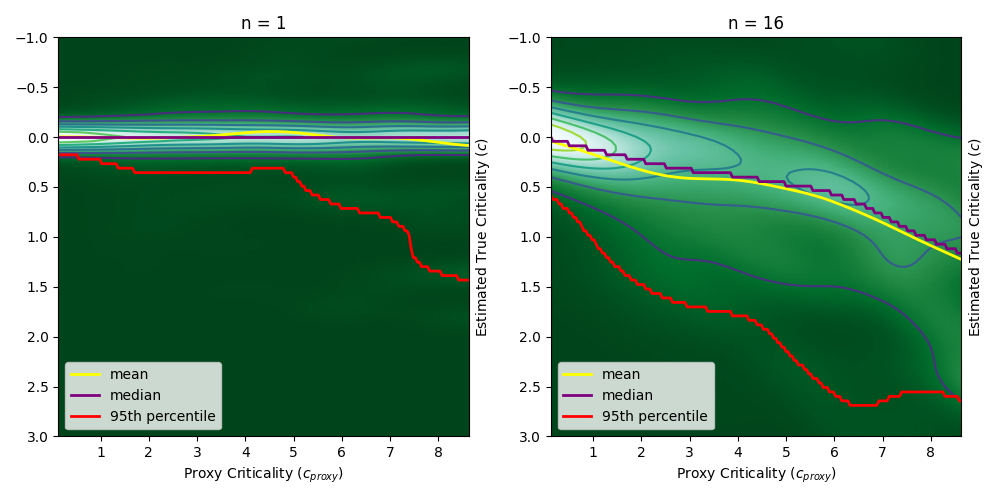}
    \caption{Normalized (over the Y-axis) kernel density plots for A3C capturing the relationship between true criticality and proxy criticality, with contours and trendlines, for 1 (left) vs. 16 (right) random actions.}
    \label{cai:fig:noise}
\end{figure}

First, we define $c_{proxy}(o_t, o_{t-1}, ...)$ as a proxy criticality metric that relies only on past or present observations $o_t, o_{t-1}, ...$, making it computable during agent execution without the need to simulate the future. Next, we define true criticality as $c(t, n; \pi) = E_{a \sim \pi}\left[R_\gamma\right] - E_{a \sim \mathcal{U}}\left[R_\gamma\right]$, the difference in expectation between the total $\gamma$-discounted reward $R_\gamma$ when actions are drawn from $\pi$, the agent's policy, versus $\mathcal{U}$, a uniform distribution, for $n$ steps starting at time $t$; before and after those $n$ steps, the episode runs as normal, drawing $a \sim \pi$. The uniform distribution is chosen rather than any adversarial policy based on $\pi$ because $\pi$ is potentially unaware of its own failings, which we are trying to quantify. To approximate true criticality, we replace the actions at times $t, ..., t+n-1$ with random actions, and measure the reduction in total discounted reward over time steps $t, ..., t+h-1$ for some large horizon $h \gg n$; this is repeated until the mean reduction in reward is $95\%$ likely to be within a small $\epsilon = 0.2$ of the true criticality \cite{driels2004}. Ideally the proxy metric is monotonic with respect to the true criticality, though often there is significant noise in this relationship; \cref{cai:fig:noise} shows kernel density plots of the empirical relationship between these, for $n = 1$ and $n = 16$.  Each such plot is generated by running $1000$ episodes, and computing the proxy and true criticality for some time $t$ in an episode, for different values $n$.  For $500$ of the episodes, $t$ is selected randomly; for the other $500$, an attempt is made to obtain a relatively uniform sample of $c_{proxy}$ values.

Next, safety margins are defined as $s(c_{proxy}(\cdot), \zeta; \pi)$, the maximum number of random actions that can be tolerated before there is an $\alpha = 0.05$ or greater chance that the sum of discounted expected future rewards will decrease by more than $\zeta$. These are computed using the $1 - \alpha$ percentile curve calculated on the density plot, as shown in \cref{cai:fig:noise}. These curves serve as region boundaries in \cref{cai:fig:results:safety}, after being adjusted to ensure monotonicity with respect to proxy metric values (with safety margins reduced when necessary for doing so).

\section*{Acknowledgments}

This material is based upon work supported by the Air Force Research Laboratory (AFRL) under Contract No. FA8750-22-C-1002.
This paper was PA approved under case number AFRL-2023-1554.
\vspace{-0.15em}

\section*{SBIR Data Rights}

Contract No.: FA8750-22-C-1002

Contractor Name: Galois, Inc.

Address: 421 SW Sixth Ave., Suite 300, Portland, OR 97204

Expiration of SBIR Data Rights Period: 01/18/2042

The Government's rights to use, modify, reproduce, release, perform, display, or disclose technical data or computer software marked with this legend are restricted during the period shown as provided in paragraph (b)(4) of the Rights in Other Than Commercial Technical Data and Computer Software - Small Business Innovation Research (SBIR) Program clause contained in the above identified contract. No restrictions apply after the expiration date shown above. Any reproduction of technical data, computer software, or portions thereof marked with this legend must also reproduce the markings.

\vspace{-0.15em}

\newcommand\lastcolumnfix{\enlargethispage{-11cm}}
\IEEEtriggercmd{\lastcolumnfix}
\bibliographystyle{sty/ieee/IEEEtran-nomonth}
\bibliography{sty/ieee/IEEEabrv,references}


\end{document}